\documentclass[conference]{IEEEtran}
\IEEEoverridecommandlockouts

\def\SM{\text{FH-TabNet}}

\usepackage{soul}
\usepackage{pifont}

\makeatletter
\newcommand{\linebreakand}{%
  \end{@IEEEauthorhalign}
  \hfill\mbox{}\par
  \mbox{}\hfill\begin{@IEEEauthorhalign}
}
\makeatother

\title{FH-TabNet: Multi-Class Familial Hypercholesterolemia Detection via a Multi-Stage Tabular Deep Learning Network\thanks{This work was partially supported by the Natural Sciences and Engineering Research Council (NSERC) of Canada through the NSERC Discovery Grant RGPIN-2023-05654.}} 

\author{\IEEEauthorblockN{Sadaf Khademi}
\IEEEauthorblockA{\textit{Concordia Ins. Inf. Syst. Eng. (CIISE)} \\
\textit{Concordia University}\\
Montreal, Canada\\
sadaf.khademi@concordia.ca}
\and
\IEEEauthorblockN{Zohreh Hajiakhondi}
\IEEEauthorblockA{\textit{Dept. of Elec. and Comp. Eng. (ECE)} \\
\textit{Concordia University}\\
Montreal, Canada\\
zohreh.hajiakhondimeybodi@concordia.ca}
\and
 
\IEEEauthorblockN{Golnaz Vaseghi}
\IEEEauthorblockA{\textit{Applied Physiology Research Center} \\
\textit{Isfahan University of Medical Sciences}\\
Isfahan, Iran \\
golnazvaseghi@yahoo.com}

\and
\linebreakand 
\IEEEauthorblockN{Nizal Sarrafzadegan}
\IEEEauthorblockA{\textit{Isfahan Cardiovascular Research Center} \\
\textit{Isfahan University of Medical Sciences}\\
Isfahan, Iran\\
nsarrafzadegan@gmail.com}
\and

\IEEEauthorblockN{Arash Mohammadi}
\IEEEauthorblockA{\textit{Concordia Ins. Inf. Syst. Eng. (CIISE)} \\
\textit{Concordia University}\\
Montreal, Canada\\
arash.mohammadi@concordia.ca}
}

\usepackage{flushend}
\usepackage{epsfig}
\usepackage{amsmath}
\usepackage{flushend}
\usepackage{multirow}
\usepackage{amsthm}
\usepackage{amssymb }
\usepackage{amsmath}
\usepackage{bm}
\usepackage{cite}
\usepackage{float}
\usepackage{mathrsfs}
\usepackage{amsfonts}
\usepackage{graphicx}
\usepackage{algorithm, algpseudocode}
\usepackage{subfigure}
\usepackage[usenames]{color}
\begin{document}

\date{\today}
\maketitle
\thispagestyle{empty}

\begin{abstract}
Familial Hypercholesterolemia (FH) is a genetic disorder characterized by elevated levels of Low-Density Lipoprotein (LDL) cholesterol or its associated genes. Early-stage and accurate categorization of FH is of significance allowing for timely interventions to mitigate the risk of life-threatening conditions. Conventional diagnosis approach, however, is complex, costly, and a challenging interpretation task even for experienced clinicians resulting in high underdiagnosis rates. Although there has been a recent surge of interest in using Machine Learning (ML) models for early FH detection, existing solutions only consider a binary classification task solely using classical ML models. Despite its significance, application of Deep Learning (DL) for FH detection is in its infancy, possibly, due to categorical nature of the underlying clinical data. The paper addresses this gap by introducing the $\SM$, which is a multi-stage tabular DL network for multi-class (Definite, Probable, Possible, and Unlikely) FH detection. The $\SM$ initially involves applying a deep tabular data learning architecture (TabNet) for primary categorization into healthy (Possible/Unlikely) and patient (Probable/Definite) classes. Subsequently, independent TabNet classifiers are applied to each subgroup, enabling refined classification. The model's performance is evaluated through $5$-fold cross-validation illustrating superior performance in categorizing FH patients, particularly in the challenging low-prevalence subcategories.
\end{abstract}
\textbf{\textit{Index Terms}---Cardiovascular Disease, Familial Hypercholesterolemia, Genetic Disorder, TabNet, Tabular Data.}
%
\section{Introduction} \label{sec:Introduction}
Familial Hypercholesterolemia (FH) is one of the most prevalent genetic disorders characterized by abnormally high levels of blood cholesterol~\cite{Arnold}. Its prevalence is estimated to range from $1$ in $200$ to $1$ in $300$ individuals across various ethnicities~\cite{Toft}. The history of FH traces its roots to the pioneering work of the Norwegian physician, Dr. Carl Müller~\cite{Muller:1938}, who shed light on the association between hypercholesterolemia and tendinous xanthomas, connecting them to cardiovascular disease through the lens of single-gene inheritance~\cite{WHO}. FH disorder is commonly caused by mutations in genes responsible for regulating cholesterol metabolism, such as the Low-Density Lipoprotein (LDL) receptor gene that can be passed down through generations in families. This genetic disorder increases the risk of early-onset cardiovascular diseases, including heart attacks and strokes, due to the excessive buildup of LDL cholesterol in the arteries~\cite{Henderson, Nordestgaard}. 
Early detection of FH is, therefore, not only cost-effective but also crucial for preserving lives. However, only $10$\% of the estimated number of worldwide affected individuals have received a formal FH diagnosis. Out of this population, only $2$\% were identified before the age of $18$ years. For the majority of those affected, FH goes unnoticed until middle ages, typically, surfacing around the age of $45$ in tandem with the development of cardiovascular disease, which highlights the urgent need for screening/diagnosis techniques at younger ages~\cite{Vallejo-Vaz, Sturm, Najam, Sadiq}. Indeed, without early recognition, many patients receive inadequate treatment and miss valuable opportunities for preventing cardiovascular problems, which can not only impact their quality of life but may also shorten their lifespan.
Despite extensive efforts in the medical community, critical challenges (outlined later in Section~\ref{Sec:RWs}) persist for early detection and timely intervention of FH. Leveraging the power of Electronic Medical Records (EMRs) and Artificial Intelligence (AI), we aim to address this gap. In this context, the paper proposes the $\SM$ framework that provides highly accurate early detection results without relying on genetic data.

\noindent
\textbf{Contributions:}
The paper introduces an innovative framework for diagnosing FH disorder in four distinct stages of progression, referred to as the Multi-Class Familial Hypercholesterolemia Detection via a Multi-Stage Tabular Deep Learning Network ($\SM$). The proposed $\SM$ framework is designed to stage individuals with FH into the following four categories: Definite, Probable, Possible, and Unlikely. A major challenge in this context is the low prevalence of certain sub-categories, which renders the use of a single-stage staging model infeasible. To address this challenge, the $\SM$ adopts a multi-stage approach, utilizing binary classification techniques built at different stages based on a tabular learning architecture known as TabNet~\cite{tabnet}. In the first stage, $\SM$ differentiates between combined Definite \& Probable category and combined Possible \& Unlikely category. In the second stage, two parallel binary classification models are designed to provide a more refined within-category assessment of the FH risk stage. Sequential attention is utilized for adaptive feature selection at each decision step, which in turn enables the underlying TabNet to more efficiently conduct end-to-end learning. In summary, the paper makes the following key contributions:
\begin{itemize}
\item Introducing the intuitively pleasing $\SM$ architecture, developed based on tabular Deep Neural Networks (DNNs). The $\SM$, to the best of our knowledge, is the first DL-based solution for multi-class FH risk categorization, providing accurate predictions for low-prevalence subcategories.
\item Providing accurate FH staging results without relying on genomic data. Incorporation of EMRs and blood markers instead of genomic data sets $\SM$ apart from its counterparts by making it more cost-effective/accessible in healthcare settings with limited resources.
\end{itemize}
Simulation results demonstrate a significant improvement in the reliability of FH risk prediction when comparing $\SM$ with traditional ML models. The $\SM$ achieved notably higher F1-scores, particularly in predicting the low-prevalence subcategory of FH patients. More specifically, through $5$-fold Cross-Validation (CV), it achieved average F1-scores of $79.20$\% for Definite, $87.20$\% for Probable, $98.60$\% for Possible, and $98.20$\% for Unlikely FH patients.

The rest of the paper is organized as follows: Section~\ref{Sec:RWs}, first, provides an overview of the relevant literature within this field. Afterwards, Section~\ref{sec:MM}, presents the data pre-processing phase and introduces TabNet as the foundational component of the $\SM$ framework. Section~\ref{SM} introduces the $\SM$ architecture. Simulation results are presented in Section~\ref{Sec:4}. Finally, Section~\ref{con} concludes the paper.


\section{Related Works}\label{Sec:RWs}

As stated previously, the term ``hypercholesterolemia" was first introduced in late $1930's$ by Carl Müller~\cite{Muller:1938} conducting a study including $17$ families in which $68$ of $76$ members showed signs of heart disease. He defined hypercholesterolemia patients with tuberous xanthomas and angina signs and concluded that this disorder is hereditary with an autosomal (a specific gene on a numbered chromosome rather than a sex chromosome) dominant characteristic.
When it comes to FH detection, Dutch Lipid Network Criteria (DLNC), Simon Broome Registrar Criteria, and Make Early Diagnosis and Prevent Early Death (MED-PED) criteria~\cite{Alonso} represent conventional FH screening methods widely employed in clinical settings for diagnosing FH. However, these established models exhibit several drawbacks affecting their practical application. The DLNC and Simon Broome criteria, incorporate lipid levels, physical examinations, family history, and when accessible, genetic data. In contrast, MED-PED criteria prioritize lipid levels and family history. The subjective nature of family history assessments, coupled with scoring variations and diagnostic threshold complexities, may lead to high levels of inconsistency. Moreover, these conventional models also face challenges related to resource accessibility, cost management, and potential population variability, highlighting the need for more effective and accessible diagnostic approaches. 

Consequently, there has been a recent surge of interest in applying Machine Learning (ML) techniques for the detection of FH. ML models have gained considerable attention in the field of medical analysis and disease detection. However, almost all of the recent research works~\cite{Myers:2019, Akyea:2020, Gratton:2023, Banda:2019, Moradifar:2023}, mainly focused on binary classification of FH based on classical and hand-crafted ML solutions. In other words, Deep Learning (DL) through the implementation of DNNs has not fully infiltrated this domain, while, traditional ML techniques have been featured in prestigious publications (i.e., Lancet)~\cite{review:2023}.
For example, Reference~\cite{Myers:2019} developed the FIND-FH model comprising two sequential layers of Random Forest (RF) models, the first one of which is used for feature selection. Incorporation of two consecutive RF layers enhanced model's performance and adaptability. Reference~\cite{Akyea:2020} evaluated the effectiveness of various conventional ML techniques in improving the detection of FH and assessed their clinical applicability within a substantial primary care patient population. Similarly, in~\cite{Gratton:2023}, a Logistic Regression (LR) model employing the Least Absolute Shrinkage and Selection Operator (LASSO) technique was utilized to discern predictive factors that effectively distinguished individuals with FH. Likewise, Reference~\cite{Banda:2019} devised a classifier using Electronic Health Record (EHR) data from Stanford Health Care to identify potential FH patients. The classifier, constructed as an RF model, underwent training on data from confirmed patients and carefully matched non-cases.  Most other recent works such as~\cite{Moradifar:2023}, followed a similar approach and focused on application of different classical ML models (i.e., RFs; Gradient Boosting; Support Vector Machine (SVM), and; LR) for the task of FH detection.
Finally, Reference~\cite{Dritsas:2022} focused on long-term risk prediction of FH by applying supervised ML models aimed at developing highly efficient risk prediction tools for FH occurrence. To identify the most effective solution, different conventional ML models such as Naive Bayes, SVM, Decision Tree, and Ensemble Learning were tested following a comprehensive hand-crafted feature analysis step.
\begin{figure*}[t!]
    \centering
    \includegraphics[scale = .10]{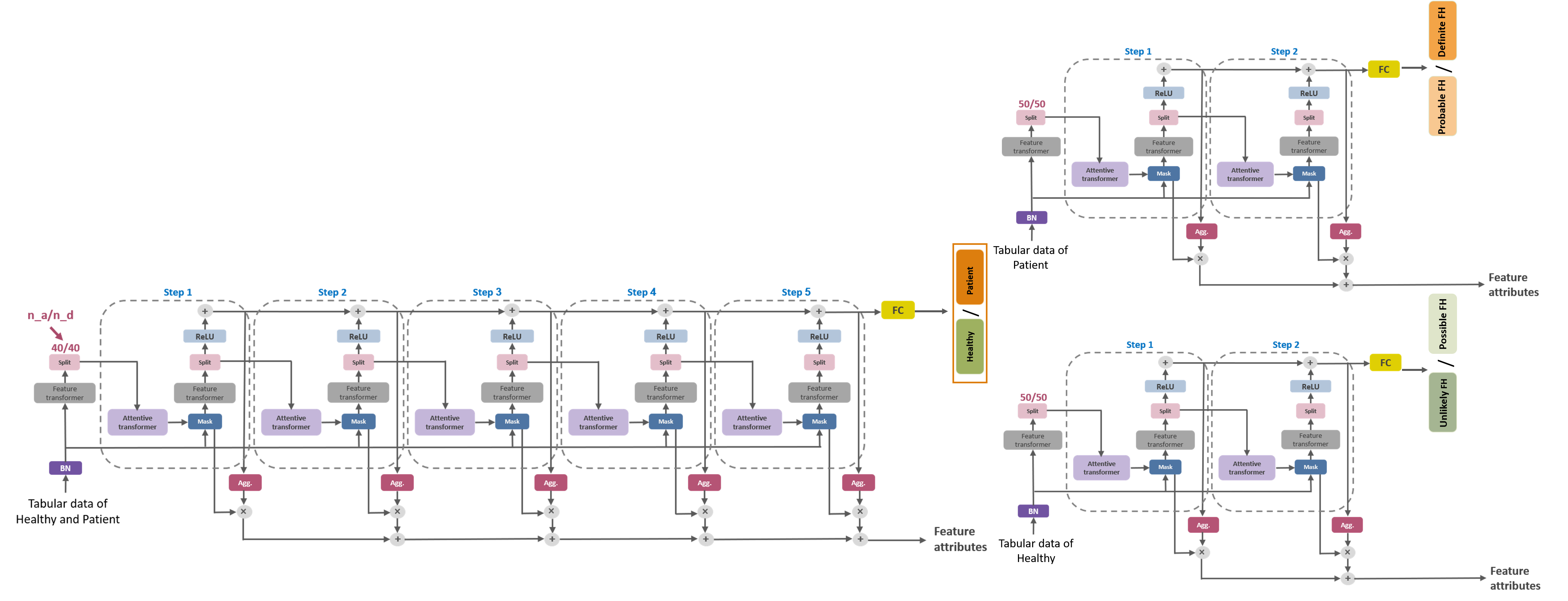}
    \caption{\footnotesize $\SM$ architecture with its building blocks. BN, Agg, and FC represent Batch Normalization, Aggregation, and Fully Connected, respectively.}
    \label{fig:tabnet}
\end{figure*}
In conclusion, while application of ML has been targeted for FH detection, to the best of our knowledge, the focus of recent research works~\cite{Myers:2019, Akyea:2020, Gratton:2023, Banda:2019, Moradifar:2023, Dritsas:2022} was restricted to development of conventional ML models considering a binary classification problem. Application of DL models within this context is in its infancy, possibly, due to the small size of available datasets and the categorical nature of the underlying clinical data. The paper aims to address this gap by targeting design of a domain-specific DL multi-class model.

\section{Materials and Methods} \label{sec:MM}


In this section, first, we will provide a concise overview of the tabular dataset employed for developing and validating the proposed $\SM$ framework. Following that, we will present the pre-processing stage, which is aimed to purify the dataset and make it appropriate for subsequent phases.

\subsection{Dataset}

This study is developed based on an in-house dataset consisting of a cohort of $1,929$ individuals with $390$ attributes after excluding patient identification details. The primary classification of subjects within this dataset relies on the FH Dutch score also known as the DLNC~\cite{Rasadi}, which was introduced as a scoring system for diagnosing FH in late $1990$s. The FH Dutch score is a reliable cumulative metric considering a range of genetic, clinical, and familial factors for determining an individual's likelihood of having FH. Based on this, individuals could be assigned to one of four distinct hypercholesterolemia classes:
\begin{itemize}
\item \textbf{\textit{Definite FH}}: Individuals in this category, typically, have high FH Dutch scores, approaching values higher than $8$.
\item\textbf{\textit{Probable FH}}: FH Dutch scores between $5$ to $8$ reflect individuals with the likelihood of having FH, but they may not meet all the criteria for a definite diagnosis.
\item\textbf{\textit{Possible FH}}: This group comprises individuals with FH Dutch scores falling in the range of $3$ to $5$ indicating the potential presence of FH. 
\item \textbf{\textit{Unlikely FH}}: Individuals in this category demonstrate FH Dutch scores approaching values lower than $3$ that make FH appear unlikely.
\end{itemize}

\subsection{ Data Pre-processing: Refinement and Cleaning}

Due to the significant proportion of missing values in the dataset, we adopt the strategy of using feature columns with missing rates below $5$\% and then excluding individuals who have missing values. The aforementioned processing steps result in creation of a more refined dataset, comprising of $1,591$ subjects and $50$ attributes. The contribution of samples within each class is as follows, Definite FH $2.9$\%, Probable FH $6.4$\%, Possible FH $40.2$\%, and Unlikely FH $50.5$\%. As can be derived from contribution values, if we merge Definite and Probable FH cases in one group and Possible and Likely FH cases in another it appears that the prevalence of individuals with FH is approximately $1$ in $10$ within the dataset. 
\subsection{Pre-processing: Categorization of Features} 

The final feature set includes a range of categorical and continuous variables, which may encompass medical records of the patient, visible evidence such as Xanthelasma and Corneal Arcus, blood-related markers such as Fasting Blood Glucose (FBG), Cholesterol levels (total, High-Density Lipoprotein (HDL), and Low-Density Lipoprotein (LDL)), and Triglycerides in addition to clinical (myocardial infarction and cardiovascular disease), and demographic (gender, age and weight) attributes. Categorical variables were processed through a one-hot encoding technique. This method involves converting categorical variables into binary vectors, where each category is represented by a binary column. 

\section{The Proposed $\SM$ Framework}\label{SM}
As stated previously, accurate identification of FH patients is essential for personalized treatment, risk assessment, genetic counseling, and resource allocation. To accurately detect FH, the proposed $\SM$ framework classifies individuals into four categories, i.e., Definite, Probable, Possible, and Unlikely FH. To this end, $\SM$ is constructed as a multi-stage and multi-step classification framework, sharing design strategies in common with the Tabular Attention Network (TabNet)~\cite{tabnet} architecture. Next, we present the overall structure and building blocks of the $\SM$ framework.

\subsection{Overall Structure of $\SM$}

As shown in Fig.~\ref{fig:tabnet}, the $\SM$ framework consists of three learning paths, each a multi-step encoder architecture, configured in two stages. Stage-$1$ of the $\SM$ framework consists of a single $5$-step encoder path distinguishing between healthy (Possible and Unlikely) and FH (Definite and Probable) patients. In Stage-$2$, which is in series with Stage-$1$ utilizing its outputs, there are two parallel $2$-step encoder paths to further refine the classification into Definite/Probable and Possible/Unlikely individuals. At each step, from the entire pool of available features in the training dataset, a subset of features is dynamically selected to make predictions. Such a feature selection approach is performed independently for each sample, in contrary to some DNNs that apply a fixed selection across the entire training dataset. Consequently, predictions for each sample are generated using distinct feature subsets, resulting in improved performance.

The number of decision steps ($n\textsubscript{step}$) plays a critical role in the model's ability to learn data representations. Hence, the $\SM$ starts with a broad overview of the data and gradually refines the representation to capture more intricate patterns. By using more decision steps in Stage-$1$, we allow the model to capture a broader range of relationships in the data, creating a foundational hierarchical representation. Then, in Stage-$2$, with fewer decision steps but a higher value of $n_a$ and $n_d$, we focus on refining the representation and making more precise distinctions between classes. Decreasing the number of decision steps in Stage-$2$ also helps to prevent overfitting due to the smaller amount of data compared to the first stage.

\subsection{Learning Pipeline of $\SM$}

At first, the data undergoes a Batch Normalization (BN) layer to replace the scaling process of the structured data. During each step, data sequentially passes through Feature Transformer and Attentive Transformer blocks:  
\begin{itemize}
\item \textit{Feature Transformer Block (FTB)}, is responsible for generating an internal feature representation. In each step, the FTB employs weights that are common across all steps, alongside weights that are learned specifically for that particular step. This approach ensures that each step's internal representation benefits from information derived from all steps. The internal representation of features is then mapped into $n_a$ and $n_d$ components through a split block. The mapped internal representation is then provided as the input to the ATB and the decision layer (ReLU activation function), respectively. 
\item \textit{The Attentive Transformer Block (ATB)}, is responsible for feature selection at each step. The ATB takes as input $n_a$ learned feature representations (obtained from the split block of the FTB) and generates a mask. This mask, which can be considered as a set of probabilities, is then used in the Mask Block to decide which features should be utilized for the current step. In other words, the feature masking layer uses the generated mask to select a subset of features by conducting an element-wise product between the original features and the generated mask. 
\end{itemize}
Output of each step is computed by applying the ReLU activation on the $n_d$ dimensioned vector obtained from the split block. Finally, outputs of all the decision steps are summed together and passed through a Fully Connected (FC) layer to classify the FH risk stage.

\section{Simulation Results} \label{Sec:4}

\begin{table}[t!]
\centering
\caption{\footnotesize 5-fold CV F1-score of the constituent parts of the proposed $\SM$ framework.}\label{tab:folds}

 \begin{tabular}{|c|c|c|c|c|c|c|}
\hline
\textbf{Stage} & \textbf{Class}  & \textbf{Fold 1} & \textbf{Fold 2} & \textbf{Fold 3}& \textbf{Fold 4}& \textbf{Fold 5}\\
\hline
\multirow{ 2}{*}{\textbf{1}} &
Patient
&  $0.91$ & $	0.93$ & $	0.93$ & $	0.98$ & $	0.92$
\\
& 
Healthy & $0.99$ & $	0.99 $ & $0.99$ & $	1$ & $	0.99$

\\
\hline
\multirow{ 2}{*}{\textbf{2(P)}} &
Probable & $0.92$ & $	0.81 $ & $0.87$ & $	0.83$ & $	0.93$
\\
& Definite
&  $0.84$ & $	0.77$ & $	0.85$ & $	0.70$ & $	0.80$
\\
\hline
\multirow{ 2}{*}{\textbf{2(H)}} &
Unlikely & $0.99$ & $	0.98 $ & $0.98$ & $	0.99$ & $	0.97$
\\
& Possible
&  $0.99$ & $	0.99$ & $	0.98$ & $	0.99$ & $	0.98$
\\
\hline
\end{tabular}
\end{table}
\begin{table*}[t!]
\centering
\caption{\footnotesize Comparison (F1-score: mean $\pm$ standard deviation) with traditional ML models based on the 5-fold CV.
}\label{tab:comp}

 \begin{tabular}{|c|c|c|c|c|}
\hline
\multirow{ 2}{*}{\textbf{Model}} & \multicolumn{4}{c|}{\textbf{F1-score}}  \\ \cline{2-5}
    & \textbf{Unlikely FH} & \textbf{Possible FH}&\textbf{Probable FH}&\textbf{Definite FH} \\
\hline
\textbf{RF}&
$97.44 \pm 0.61$ & $96.14 \pm 0.82$ & $	72.86 \pm 8.09	 $ & $34.66 \pm 9.62$
\\
\hline 		
\textbf{LDA}&
$ 93.94 \pm 0.84 $ & $93.70 \pm 1.43 $ & $	72.38 \pm 6.32 $ & $65.54 \pm 12.28 $
\\
\hline
\textbf{LR}&
$93.12 \pm 2.41$ & $90.20 \pm 1.38 $ & $	33.44 \pm 9.34 	$ & $19.64 \pm 18.99	$
\\
\hline
\textbf{Ridge}&
$86.52 \pm 2.76$ & $83.96 \pm 1.55$ & $	0 $ & $58.32 \pm 10.74 $
\\
\hline
\textbf{ADA}&
$86.36 \pm 10.28$ & $77.82 \pm 14.63	$ & $	17.30 \pm 12.96	 $ & $20.00 \pm 21.72$
\\
\hline
\textbf{KNN}&
$60.98 \pm 5.45$ & $65.92 \pm 4.61	$ & $	34.60 \pm 27.85 $ & $17.94 \pm 17.49$
\\
\hline
\textbf{Single-Stage Multi-Class TabNet}&
$94.20 \pm 1.93$ & $95.60 \pm 2.15$ & $	64.40 \pm 7.68 	$ & $ 56.20 \pm 12.28 $
\\
\hline
\textbf{ Proposed $\SM$ }&
$\boldsymbol{98.20 \pm 0.74}$ & $\boldsymbol{98.60 \pm 0.48}	$ & $	\boldsymbol{87.20 \pm 4.74 }	$ & $\boldsymbol{79.20 \pm 5.41} $
\\
\hline
\end{tabular}
\end{table*}
In this section, we evaluate performance of the proposed $\SM$ architecture through a $5$-fold CV experiment. Given the imbalanced nature of the data, we employ stratified CV to guarantee that each fold retains the same class distribution as the original dataset. In our experiment, $\SM$ underwent training for $250$ epochs, with a step learning rate decay strategy starting from $9e-2$ with a step size of $50$ and a decay factor of $0.9$. Both stages have identical values for $n_a$ and $n_d$, with values of $40$ and $50$, respectively, for the first and second stages.   
The F1-score is utilized as the primary metric due to data's inherent imbalance, allowing to better showcase the model's efficacy within this specific context. Table~\ref{tab:folds} presents the $5$-fold CV results of the proposed $\SM$ framework, encompassing Stage-$1$, Stage-$2$(P), and Stage-$2$(H). Performance of Stage-$1$ is indeed commendable, as evidenced by the F1-score reported in Table~\ref{tab:folds}. Furthermore,  Stage-$1$ achieved Accuracy of $98.8$\%, Sensitivity of $99.6$\%, and Specificity of $91.4$. We can conclude that Stage-$1$ effectively distinguishes between two categories, i.e., Healthy (``Unlikely'' and ``Possible''), and Patient (``Probable'' and ``Definite''). Healthy Class constitutes approximately $90.7\%$ of the dataset, while Patient Class accounts for approximately $9.3\%$. This underscores Stage-$1$'s impressive performance, illustrating its ability to effectively handle and classify data even in the face of such class distribution disparities. 
Furthermore, when we examine the F1-scores across the five folds in Stage-$1$, we observe consistently high values, ranging from $0.91$ to $1$. These results demonstrate the model's ability to perform well across different subsets of the data, further validating its effectiveness in classifying the distinct categories. 
The predicted classes by Stage-$1$ are given to the parallel binary classification models, where Stage-$2$(H) distinguishes between the ```Unlikely'' and ``Possible'' FH categories. Remarkably, owing to the substantial dataset presence in this category, this stage exhibits outstanding performance, as indicated by F1-scores ranging from $0.97$ to $0.99$ across various folds. Finally, Stage-$2$(P) classifies ``Probable'' and ``Definite'' FH patients. Considering the fact that a small portion of data lies in this classification, the F1-score of the Stage-$2$(P) classifier is less than the Stage-$2$(H). As a final note, we report AUC values associated with the $\SM$. For Stage-$2$(H), the AUC values are $0.99$, $0.99$, $0.99$, $0.99$, and $0.99$ and for Stage-$2$(P), the values are $0.95$, $0.87$, $0.84$, $0.91$, and $0.85$, respectfully for folds $1$-$5$. 

Additionally, we compare the performance of the proposed $\SM$ framework particularly in terms of the average F1-score across $5$-fold CV, with several established traditional ML models and a single-stage multi-class TabNet framework. These models encompass RF, LR, Linear Discriminant Analysis (LDA), Ridge, AdaBoost, and K-Nearest Neighbors (KNN) classifiers. According to the information provided in Table~\ref{tab:comp}, it is evident how well the models perform in predicting the ``Unlikely ``and ``Possible'' categories, which are the dominant categories due to their higher sample count. It should be noted that all models, including the $\SM$, excel in predicting ``Unlikely'' and ``Possible'' FH classes. However, what sets the $\SM$ framework apart is its exceptional performance in predicting the ``Probable'' and ``Definite'' classes. The primary reason behind this remarkable performance lies in the multi-step binary classification approach adopted within the $\SM$ framework dealing with tabular data. This innovative approach results in a more balanced distribution of data in both Stage-$2$(P) and Stage-$2$(H), which, in turn, leads to significantly improved multi-class predictions, as evidenced by the substantially higher F1-scores in all classes. Finally, we analyzed high-weight features by aggregating their scores and ranking them to identify the most pivotal features in classifying sub-categories of FH. Our findings reveal that LDL-Code-FH, cardiovascular disease, or a history of diagnosis stand out as the most important features in classifying ``Possible'' and ``Unlikely'', supported by their respective average scores of $0.33$, $0.13$, and $0.11$ across different folds. Similarly, in classification of ``Definite'' and ``Probable'', we identified the pivotal features as Corneal Arcus, history of cancer in the family, and LDL-Code-FH, which exhibit importance average weight scores of $0.25$, $0.16$, and $0.08$, respectively. These scores signify the aggregated importance of these features, showcasing their consistent significance in the classification process across multiple evaluation instances.

\section{Conclusion}\label{con}

The paper introduced an innovative classification framework, referred to as the $\SM$, designed specifically for tabular data related to a genetic disorder known as Familial Hypercholesterolemia (FH), which is characterized by high levels of LDL cholesterol in the blood leading to an increased risk of cardiovascular problems. The $\SM$ framework aims to provide precise predictions regarding the risk stage of FH (Definite, Probable, Possible, and Unlikely FH). The FH-TabNet model, as proposed, accomplishes this goal through a multi-stage application of binary classification methods. Initially, we streamlined the task by merging the Definite and Probable classes into one category and the Possible and Unlikely classes into another, thereby transforming the problem into binary classification, distinguishing between FH and healthy patients. Following this, we employed two parallel binary classification models for each subcategory, leading to a more detailed refinement of the FH risk stage. The FH-TabNet has notably enhanced the dependability of FH risk prediction when contrasted with conventional ML models. It has achieved notably higher F1-scores, particularly in the prediction of the low-prevalence subcategory of FH patients.

\bibliographystyle{IEEEbib}

\end{document}